\documentclass{article}
\pdfoutput=1
\usepackage{spconf,amsmath,graphicx}

\usepackage{hyperref}
\usepackage{url}

\usepackage[utf8]{inputenc} 
\usepackage[T1]{fontenc}    
\usepackage{booktabs}       
\usepackage{amsfonts}       
\usepackage{nicefrac}       
\usepackage{microtype}      
\usepackage{subcaption}
\usepackage{paralist}
\usepackage{floatrow}
\newfloatcommand{capbtabbox}{table}

\usepackage{array}
\newcolumntype{T}{>{\centering\arraybackslash}m{0.78cm}}
\newcolumntype{Z}{>{\raggedright\arraybackslash}m{0.7cm}}
\newcolumntype{X}{>{\raggedright\arraybackslash}m{2.0cm}}
\newcolumntype{Y}{>{\centering\arraybackslash}m{2.0cm}}
\newcolumntype{M}[1]{>{\centering\arraybackslash}m{#1}}

\usepackage{xcolor}

\usepackage{graphicx}
\usepackage{setspace}

\newcommand{\capspace}{\vspace{-0.5\baselineskip}}

\title{
Deep Metric Learning and Image Classification \\ with  Nearest Neighbour Gaussian Kernels
}

\name{Benjamin J. Meyer, Ben Harwood, Tom Drummond\vspace{-.5cm}\thanks{This research was supported by the Australian Research Council Centre of Excellence for Robotic Vision (project number CE140100016).}}
\address{ARC Centre of Excellence for Robotic Vision, Monash University\vspace{-.04cm} \\
 \texttt{\{benjamin.meyer,ben.harwood,tom.drummond\}@monash.edu}
 }

\begin{document}
%
\maketitle

\begin{abstract}
We present a Gaussian kernel loss function and training algorithm for convolutional neural networks that can be directly applied to both distance metric learning and image classification problems. Our method treats all training features from a deep neural network as Gaussian kernel centres and computes loss by summing the influence of a feature's nearby centres in the feature embedding space. Our approach is made scalable by treating it as an approximate nearest neighbour search problem. We show how to make end-to-end learning feasible, resulting in a well formed embedding space, in which semantically related instances are likely to be located near one another, regardless of whether or not the network was trained on those classes. Our approach outperforms state-of-the-art deep metric learning approaches on embedding learning challenges, as well as conventional softmax classification on several datasets.
\end{abstract}

\section{Introduction} \label{introduction}

Metric learning aims to learn a transformation from the image space to a feature embedding space, in which distance is a measure of semantic similarity. Feature embeddings from semantically similar images will be located nearby, while those of semantically dissimilar images will be located far apart. Applications for such effective feature embeddings include transfer learning, retrieval, zero- and few-shot learning, clustering and weakly or self supervised learning. 
Image classification is the task of categorising an image into one of a set of classes. Applications include object  
and scene recognition.

Classification and metric learning are generally treated as separate problems. As such, metric learning approaches have struggled to reach the classification performance of state-of-the-art classifiers.
Likewise, classification approaches fail to learn feature spaces that represent inter- and intra-class similarities to the standard of metric learning approaches.
Outside of zero- and few-shot learning, some metric learning algorithms have been applied to classification \cite{rippel2016metric,NIPS2016_6200}, although approaches that perform well in both domains remain uncommon.

We propose a novel loss function and training algorithm for convolutional neural networks that can be applied to both metric learning and image classification problems, outperforming conventional approaches in both domains. Our approach defines training set feature embeddings as Gaussian kernel centres, which are used to push or pull features in a local neighbourhood, depending on the labels of the associated training examples.
Fast approximate nearest neighbour search is used to provide an efficient and scalable solution. Our approach differs from kernel or radial basis function neurons \cite{broomhead1988radial,xu2001kernel}, as our kernel centres are not learned network parameters, but are defined to be the locations of the training set features in the embedding space.
Additionally, we use kernels only in the loss function and classifier, not as activation functions throughout the network. Beyond activation functions, kernels have also been used in neural network classifiers as support vector machines \cite{Razavian:2014:CFO:2679599.2679731,donahue2014decaf,tang2013deep}. Our approach is related to NCA \cite{goldberger2005neighbourhood,salakhutdinov2007learning}, but introduces per exemplar weights, makes training feasible through the introduction of periodic asynchronous updates of the kernel centres, and is made scalable for a large number of training examples and a high embedding dimension. Additionally, we explore the importance of the embedding space dimensionality.

The best success on embedding learning tasks has been achieved by deep metric learning methods \cite{DBLP:journals/corr/HofferA14,7298682,7780803,NIPS2016_6200,1704.01285}, which make use of deep neural networks. The majority of these approaches use or generalise a triplet architecture with hinge loss \cite{weinberger2006distance}, although including global loss terms can also be beneficial \cite{DBLP:journals/corr/SongJR016,7780950,1704.01285}. 
Triplet networks take a trio of inputs; an anchor image, an image of the same class as the anchor and an image of a different class.
Triplet approaches aim to map the anchor nearer the positive example than the negative example, in the feature space. 
Such approaches may indiscriminately pull examples of the same class together, regardless of the local structure of the space. In other words, these methods aim to form a single cluster per class, limiting the intra- and inter-class similarities that can be represented. In contrast, our approach considers only the local neighbourhood of a feature, allowing multiple clusters to form for a single class, if that is appropriate. 
Our approach outperforms state-of-the-art deep metric learning approaches on embedding learning challenges.

The most common approach to image classification is a convolutional neural network trained with softmax loss, which transforms activations into a distribution across class labels \cite{NIPS2012_4824,Simonyan14c,7298594,7780459}. However, softmax is inflexible as classes must be axis-aligned and the number of classes is baked into the network. Our approach is free to position clusters such that the intrinsic structure of the data can be better represented.
Our metric learning approach to classification outperforms softmax on several datasets, while simultaneously representing the intra- and inter-class similarities sought by metric learning approaches.
Metric learning also allows for new classes to be added on-the-fly, with no updates to the network weights required to obtain reasonable results. 

The advantages of our approach are as follows:
\begin{compactitem}
	\item Training is made feasible by introducing periodic asynchronous updates of the kernel centres (Section \ref{end_to_end}).
	\item End-to-end learning can be made scalable by leveraging fast approximate nearest neighbour search (Section \ref{nearest_neighbour}).
	\item Our approach can be applied to two separate problems; image classification and metric learning.
	\item Our approach outperforms state-of-the-art deep metric learning algorithms on the Stanford Cars196 and CUB Birds200 2011 datasets (Section \ref{metric_results}).
	\item Finally, our approach outperforms a conventional softmax classifier on the fine-grained classification datasets CUB Birds200 2011, Stanford Cars196, Oxford 102 Flowers and Leafsnap (Section \ref{classification_results}).
\end{compactitem}

\section{Approach} \label{method}

\begin{figure}[!t]
	\centering
	\includegraphics[width=\linewidth]{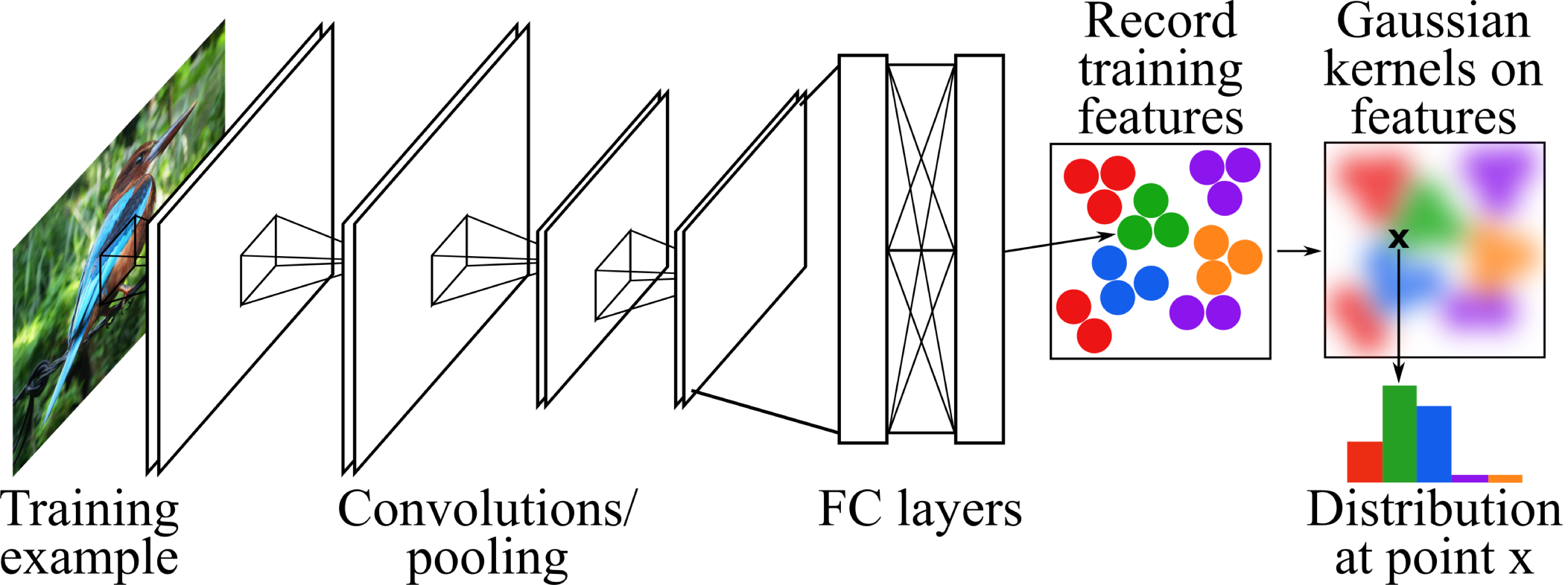}
	\vspace{-1.5\baselineskip}
	\caption{Overview of our approach. Note that the feature embeddings are high dimensional.
	}
	\label{overview}
\end{figure}

A Gaussian kernel returns a value that depends only on the distance between a point $\mathbf{x}$ and the Gaussian centre $\mathbf{c}$. The Gaussian kernel $f$ is calculated as:
\begin{equation} \label{kernel_eq}
f(\mathbf{x},\mathbf{c}) = \exp \left ( \frac{-\| \mathbf{x} - \mathbf{c} \| ^ 2}{2 \sigma ^2} \right ),
\end{equation}
where $\sigma$ sets the kernel width. We define the feature embeddings of each training set example as Gaussian kernel centres. Specifically, in a deep neural network we take the layer immediately before the loss function or classifier as the embedding layer. For example, in a VGG architecture, this may be FC7 (fully connected layer 7), forming a 4096 dimension embedding. In general, however, the embedding may be of any size. An overview of this approach is seen in Figure \ref{overview}.

\subsection{Classifier and Loss Function}

A classifier is formed by the weighted sum of the kernel distance calculations between a feature embedding and the centres. Classification of an example is achieved by passing the input through the network, resulting in a feature embedding in the same space as the centres. A probability distribution over class labels is found by summing the influence of each centre and normalising. A centre contributes only to the class of the training example coupled to that centre. For example, the probability that feature embedding $\mathbf{x}$ has class label $Q$ is:
\begin{equation} \label{prob_RBF}
Pr(\mathbf{x} \in \mbox{class }Q) = \frac{ \sum_{i \in Q} w_i f(\mathbf{x},\mathbf{c_i})} { \sum_{j=1}^{m} w_j f(\mathbf{x},\mathbf{c_j}) },
\end{equation}
where $f$ is the kernel, $i \in Q$ are the centres with label $Q$, $m$ is the number of training examples and $w_i$ is a weight for centre $i$, which is learned end-to-end with the network weights. Note that a global $\sigma$ value is shared by all kernels. If an example is in the training set, the distance calculation to itself is omitted during the computation of the classification distribution, the loss function and the derivatives.

The loss function used is the summed negative logarithm of the probabilities of the true class labels. For example, the loss for example $\mathbf{x}$ with ground truth label $R$ is
$-\ln \left(Pr(\mathbf{x} \in \mbox{class }R) \right)$.
The same loss function is used for both classification and metric learning problems.

\subsection{Nearest Neighbour Gaussian Kernels} \label{nearest_neighbour}

Equation \ref{prob_RBF} is calculated by summing over all kernels. However, since the centres are attached to training examples, of which there can be any large number, computing that sum is both intractable and unnecessary. Most kernel values for a given example will be effectively zero, as the feature will lie only within a subset of the Gaussian windows. As such, we consider only the local neighbourhood of a feature embedding. Considering the nearest Gaussian centres to a feature ensures that most of the distance computations are pertinent to the loss calculation. The classifier equation becomes:
\begin{equation} \label{prob_nnRBF}
Pr(\mathbf{x} \in \mbox{class }Q) = \frac{ \sum_{i \in Q \cap \mathcal{N}} w_i f(\mathbf{x},\mathbf{c_i})} { \sum_{j \in \mathcal{N}} w_j f(\mathbf{x},\mathbf{c_j}) },
\end{equation}
where $\mathcal{N}$ is the set of approximate nearest neighbours for example $\mathbf{x}$ and $i \in Q \cap \mathcal{N}$ is the set of approximate nearest neighbours that have label $Q$. Again, training set examples exclude their own centre from their nearest neighbour list.

In the interest of providing a scalable solution, we use approximate nearest neighbour search to obtain candidate nearest neighbour lists. This allows for a trade off between precision and computational efficiency. Specifically, we use a Fast Approximate Nearest Neighbour Graph (FANNG) \cite{7780985}, as it provides the most efficiency when needing a high probability of finding the true nearest neighbours of a query point. Importantly, FANNG provides scalability in terms of the number of dimensions and the number of training examples.

\subsection{Training the Network} \label{end_to_end} 

The Gaussian centre locations change as the network weights are updated each training iteration. Although required to compute the derivatives, it is intractable to find the true locations of each example's neighbouring centres online during training. However, we find that it is not necessary for the centres to be up to date at all times in order for the model to converge. We store a bank of the Gaussian centres and perform periodic asynchronous updates of all centres at a fixed interval.

As the centres change, the nearest neighbours also change. Again, it is intractable to compute the correct nearest neighbours each time the network weights are updated. This is remedied by considering a larger number of nearest neighbours than would be required if all centres and neighbour lists were up-to-date at all times. The embedding space changes slowly enough that it is highly likely many of the previously neighbouring centres will remain relevant. Since the Gaussian kernel decays to zero as the distance between the points becomes large, it does not matter if a centre that is no longer near the example remains a candidate nearest neighbour. 

We call the interval at which the Gaussian centres are updated and the nearest neighbours computed during training the \textit{update interval}. 
This interval is training set dependant and we find intervals between 1 and 10 epochs work well in our experiments. Note that the stored Gaussian centres do not have dropout \cite{srivastava2014dropout} applied, but the current training embeddings may.

\section{Experiments} \label{results}

\subsection{Distance Metric Learning} \label{metric_results}

We evaluate our approach on Stanford Cars196 (16,185 images of 196 car models) \cite{krause20133d} and CUB Birds200 2011 (11,788 images of 200 bird species) \cite{WelinderEtal2010}. In this problem, the network is trained and evaluated on different sets of classes. Following the set-up in \cite{7780803,NIPS2016_6200,DBLP:journals/corr/SongJR016,1704.01285}, we train on the first half of classes and evaluate on the remaining classes. Stochastic gradient descent optimisation is used. Images are resized to 256x256 and data is augmented by random cropping and horizontal mirroring. 
The object bounding boxes are not used. GoogLeNet \cite{7298594} with ImageNet \cite{russakovsky2015imagenet} pre-trained weights is used as the model. We use 100 nearest neighbours, an update interval of 10 epochs, batch size of 20, base learning of 
0.00001 and weight decay of 0.0002. The Gaussian $\sigma$ used depends on the number of dimensions of the feature embedding; values between 10 and 30 work well for this task. We evaluate on two metrics. The first, Normalised Mutual Information (NMI) \cite{manning2008introduction}, is a clustering metric that finds the ratio of mutual information and average entropy of a set of clusters and labels. The second, Recall@K (R@K), defines a true positive as an example feature embedding that has at least one out of its true K nearest neighbours with the same class as itself.

\begin{figure}[!t]
\captionsetup[subfigure]{aboveskip=-1pt,belowskip=-1pt}
\centering
    \begin{subfigure}{0.5\linewidth}
    	\includegraphics[width=\linewidth]{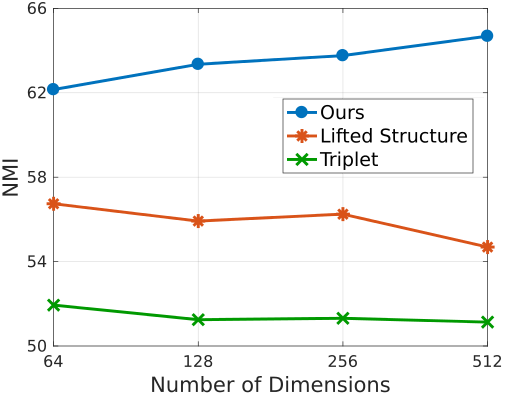}
    	\caption{}
    	\label{nmi_cars_plot}
    \end{subfigure}%
    \begin{subfigure}{0.5\linewidth}
    	\includegraphics[width=\linewidth]{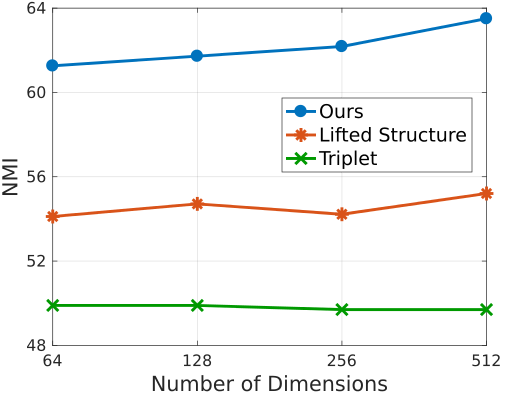}
    	\caption{}
    	\label{nmi_birds_plot}
    \end{subfigure}
    \begin{subfigure}{0.5\linewidth}
    	\includegraphics[width=\linewidth]{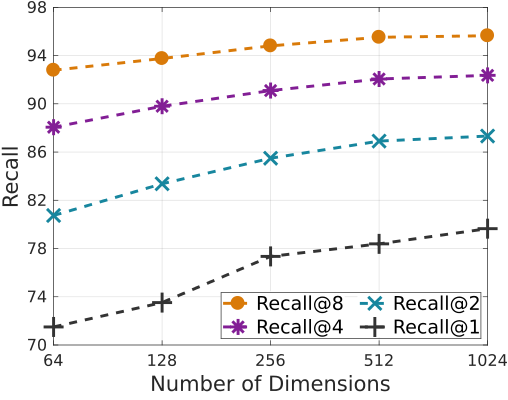}
    	\caption{}
    	\label{recall_cars_plot}
    \end{subfigure}%
    \begin{subfigure}{0.5\linewidth}
    	\includegraphics[width=\linewidth]{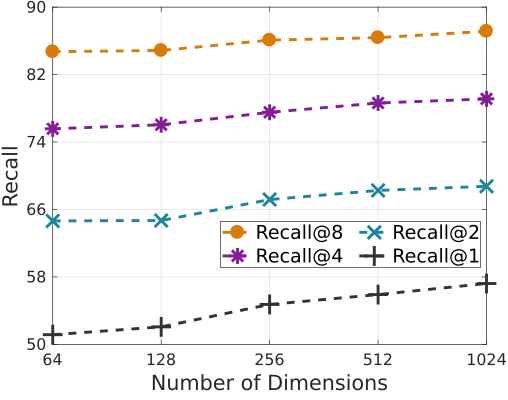}
    	\caption{}
    	\label{recall_birds_plot}
    \end{subfigure}
    \capspace
    \caption{Top: NMI score with increasing embedding dimension on Cars196 (a) and Birds200 (b). Bottom: Recall at $K$ nearest neighbours for our approach on Cars196 (c) and Birds200 (d).
    }
\end{figure}

We first investigate the importance of the feature embedding dimension, which is set by the output dimensionality of a final fully connected layer. 
A similar study in \cite{7780803} suggests that the number of dimensions is not important for triplet networks, in fact, increasing the number of dimensions can be detrimental to performance. We compare our method with increasing dimension size against triplet loss \cite{weinberger2006distance,7298682} and lifted structured embedding \cite{7780803}, both taken from the study in \cite{7780803}. Figures \ref{nmi_cars_plot} and \ref{nmi_birds_plot} show the effect of the embedding size on NMI score. While increasing the number of dimensions does not necessarily improve performance for triplet-based networks, higher dimensionality can be utilised by our approach, as the NMI score improves as the dimensionality increases. Similar behaviour is seen in Figures \ref{recall_cars_plot} and \ref{recall_birds_plot}, which show the Recall@K metric for our approach. Again, this shows that our approach can take advantage of a higher dimensionality.

Our approach is compared to the state-of-the-art in Tables \ref{metric_table_cars} and \ref{metric_table_birds}, with the compared results taken from \cite{DBLP:journals/corr/SongJR016} and \cite{1704.01285}. Since, as discussed above, the dimensionality does not have much impact on the other approaches, all results in \cite{DBLP:journals/corr/SongJR016} and \cite{1704.01285} are reported using 64 dimensions. For fair comparison, we report our results at 64 dimensions, but also at the better performing higher dimensions. Our approach outperforms the other methods in both the NMI and Recall@K measures, at all embedding sizes presented. Our approach is able to produce better compact embeddings than existing methods, but can also take advantage of a larger embedding space.

\subsection{Image Classification} \label{classification_results}

We compare classification performance with conventional softmax loss. Images are resized to 256x256 and random cropping and horizontal mirroring is used for data augmentation. Unlike in Section \ref{metric_results}, we crop Birds200 and Cars196 images using the provided bounding boxes before resizing. The same classes are used for training and testing. All datasets are split into training, validation and test sets. 
For all approaches,
we select hyperparameters that minimise the validation loss. For our approach with a VGG \cite{Simonyan14c} or AlexNet \cite{NIPS2012_4824} architecture, the FC7 layer (4096 dimensions), with dropout and without a ReLU, is used as the embedding layer. For a ResNet architecture \cite{7780459}, we use the final pooling layer (2048 dimensions). We find that following the ResNet embedding layer with a dropout layer results in a small performance gain for both our approach and softmax. A batch size of 20, update interval of 10 epochs and base learning rate of 0.00001 are used for our approach. We use stochastic gradient descent optimisation. A Gaussian $\sigma$ of around 100 is found to be suitable for the 4096 dimension VGG16 embeddings on Birds200. Networks are initialised with ImageNet \cite{russakovsky2015imagenet} pre-trained weights.

\begin{table}[t]
\centering
\setlength\tabcolsep{4.0pt}
\begin{tabular}{p{2.1cm} @{\hskip1pt} TTTTTT}
\toprule 
& Dims & R@1 & R@2 & R@4 & R@8 & NMI \\
\midrule
Semi-hard \cite{7298682}
	&	64	&	51.54	&	63.78	&	73.52	&	82.41	&	53.35 \\
LiftStruct \cite{7780803}
	&	64	&	52.98	&	65.70	&	76.01	&	84.27	&	56.88 \\
N-pairs \cite{NIPS2016_6200}
	&	64	&	53.90	&	66.76	&	77.75	&		86.35	&	57.79 \\
Tripl/Gbl \cite{7780950}
	&	64	&	61.41	&	72.51	&	81.75	&	88.39	&	 58.20 \\
Cluster \cite{DBLP:journals/corr/SongJR016}
	&	64	&	58.11	&	70.64	&	80.27	&		87.81	& 59.04 \\
SmrtMine \cite{1704.01285}
 	&		64	&	64.65	&	76.20	&	84.23&  90.19	& 59.50 \\
Ours	&	64	&	\textbf{71.05}	&	\textbf{80.74}	&	\textbf{88.06}	&		\textbf{92.79}	&	 \textbf{62.15} \\
\midrule
Ours	&	128	&	73.52	&	83.37	&	89.80	&		93.76	&	 63.35 \\
Ours	&	256	&	77.35	&	85.49	&	91.10	&		94.81	&	 63.76 \\
Ours	&	512	&	78.39	&	86.91	&	92.06	&		95.52	&	 64.68 \\
Ours	&	1024	&	\textbf{79.65}	&	\textbf{87.33}	&	\textbf{92.36}	&	\textbf{95.65}	& \textbf{65.30} \\
\bottomrule
\end{tabular}
\capspace
\caption{Embedding results on Cars196.}
\label{metric_table_cars}
\end{table}

\begin{table}[t]
\centering
\setlength\tabcolsep{4.0pt}
\begin{tabular}{p{2.1cm} @{\hskip1pt} TTTTTT}
\toprule 
& Dims & R@1 & R@2 & R@4 & R@8 & NMI \\
\midrule
Semi-hard \cite{7298682}
	&	64 &	42.59 &	55.03 &	66.44 &	77.23	&	55.38 \\
LiftStruct \cite{7780803}
	&	64	 &	43.57 &	56.55 &	68.59 &	79.63&	56.50 \\
N-pairs \cite{NIPS2016_6200}
	&	64 &	45.37 &	58.41 &	69.51 &	79.49	&	57.24 \\
Tripl/Gbl \cite{7780950}
	&	64 &	49.04 &	60.97 &	72.33 &	81.85	&	 58.61 \\
Cluster \cite{DBLP:journals/corr/SongJR016}
	&	64 &	48.18 &	61.44 &	71.83 &	81.92	& 59.23 \\
SmrtMine \cite{1704.01285}
 	&		64 &	49.78 &	62.34 &	74.05 &	83.31	& 59.90 \\
Ours	&	64 &	\textbf{51.15} & \textbf{64.64} & \textbf{75.57} &	\textbf{84.72}	&	 \textbf{61.26} \\
\midrule
Ours	&	128 &	52.08 &	64.69 &	76.05 &	84.86	&	 61.72 \\
Ours	&	256 &	54.74 &	67.18 &	77.53 &	86.09	&	 62.18 \\
Ours	&	512 &	55.91 &	68.26 &	78.63 &	86.38	&	 63.50 \\
Ours	&	1024 &	\textbf{57.22} &	\textbf{68.75} &	\textbf{79.12} &	\textbf{87.14}	& \textbf{63.95} \\
\bottomrule
\end{tabular}
\capspace
\caption{Embedding results on Birds200.}
\label{metric_table_birds}
\end{table}

\begin{figure}[!t]
	\centering
	\includegraphics[width=0.6\linewidth]{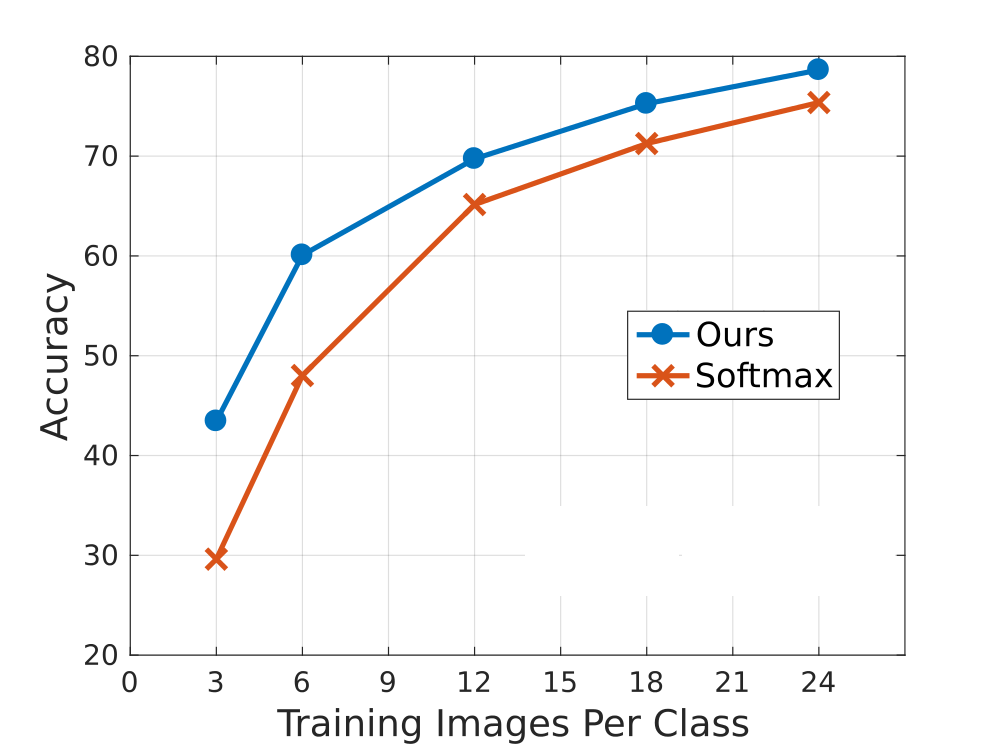}
	\capspace
	\caption{Effect of the number of training examples per class on the test set accuracy of Birds200, with a VGG16 architecture.
	}
	\label{num_training_samples}
\end{figure}

\begin{table}[t]
\centering
\begin{tabular}{@{}XYY@{}}
\toprule 
Base Network & Softmax & Ours \\
\midrule
AlexNet & 62.41 & \textbf{66.95} \\
VGG16 & 75.37 & \textbf{78.63} \\
ResNet50 & 78.05 & \textbf{78.98} \\

\bottomrule
\end{tabular}
\capspace
\caption{Birds200 test set accuracy with various architectures.}
\label{cub_table}
\end{table}

\begin{table}[!h]
\centering
\begin{tabular}{p{3cm}M{2cm}M{2cm}}
\toprule 
 Dataset & Softmax & Ours \\
\midrule
Oxford 102 Flowers & 82.79 & \textbf{86.26} \\
Stanford Cars196 & 85.67 & \textbf{86.52} \\
Leafsnap Field & 73.80 & \textbf{75.96} \\
\bottomrule
\end{tabular}
\capspace
\caption{Test accuracy on fine-grained classification datasets.}
\label{other_datasets}
\end{table}

We first evaluate on the Birds200 dataset. Since there is no standard validation set for this dataset, we take 20\% of the training data as validation data. In Table \ref{cub_table}, we evaluate with three network architectures; AlexNet \cite{NIPS2012_4824}, VGG16 \cite{Simonyan14c} and ResNet50 \cite{7780459}.
Additionally, the effect of the number of training examples per class is shown in Figure \ref{num_training_samples}. Our approach outperforms softmax loss at all numbers of training images, with a particularly large gain when training data is scarce. Further, we investigate the importance of the per kernel weights, e.g. $w_i$ from Equation \ref{prob_nnRBF}, and find that learning the weights end-to-end with the network results in a 0.69\% increase in accuracy, compared with fixing the weights at a value of one. 

We further evaluate our approach on three other fine-grained classification datasets; Oxford 102 Flowers \cite{Nilsback08}, Stanford Cars196 \cite{krause20133d} and Leafsnap \cite{leafsnap_eccv2012}. We use the standard training, validation and test splits for Oxford 102 Flowers. For Stanford Cars196, we take 30\% of the training set as validation data. We use the challenging \textit{field} images from Leafsnap, which are taken in uncontrolled conditions. The dataset contains 185 classes of leaf species and we split the data into 50\%, 20\% and 30\% for training, validation and testing, respectively. Results are shown in Table \ref{other_datasets}.

\section{Conclusion}
Our novel nearest neighbour Gaussian kernel approach to deep metric learning outperforms state-of-the-art metric learning approaches on embedding learning problems. Additionally, our approach is able to outperform conventional softmax loss when applied to classification problems. Importantly, the same loss function and training algorithm is used for both of these target domains.

\bibliographystyle{IEEEbib}
\setstretch{0.92}

\begin{table*}[!h]
\caption{Ablation study for classification on Birds200.}
\centering
\begin{tabular}{M{1.5cm}M{1.5cm}M{1.5cm}M{1.5cm}M{1.5cm}}
\toprule 
 Initial Network Weights & Tune $\sigma$ & Learn RBF Weights & Fine-tune Network Weights & Test Accuracy \\
\midrule
Random & Yes & No & No & 1.35 \\
ImageNet & Yes & No & No & 47.32 \\
ImageNet & Yes & Yes & No & 49.22 \\
ImageNet & Yes & No & Yes & 77.94 \\
ImageNet & Yes & Yes & Yes & 78.63 \\
\bottomrule
\end{tabular}
\label{ablation}
\end{table*}

\begin{figure*}[!h]
\begin{subfigure}{.3333\textwidth}
  \centering
  \includegraphics[width=\linewidth]{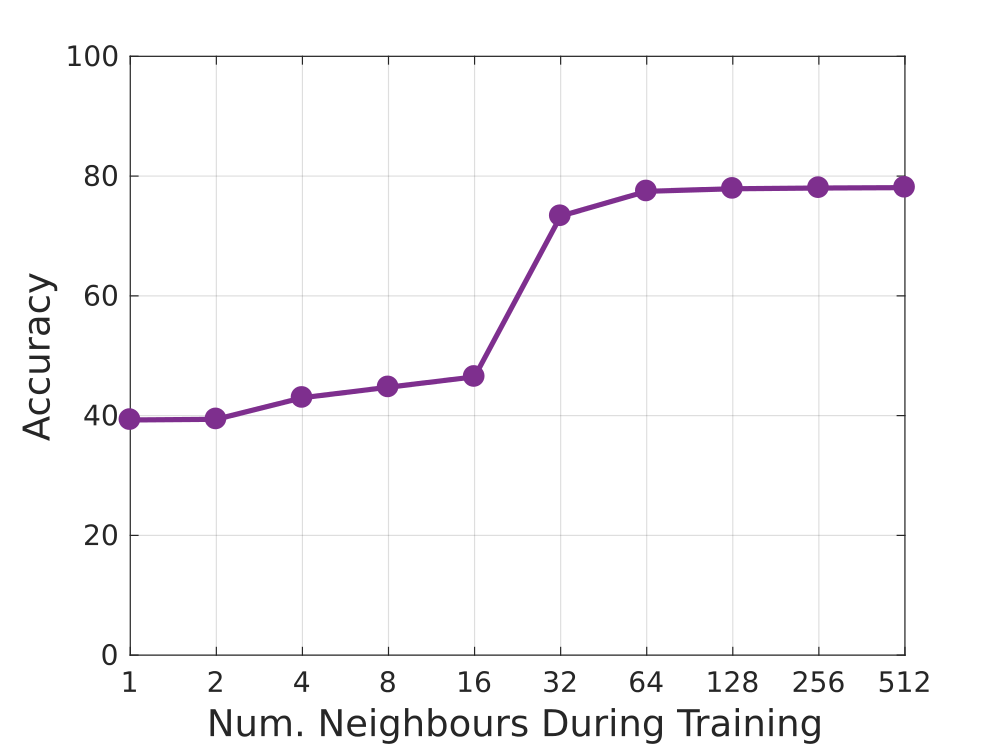}
  \caption{}
  \label{num_nn}
\end{subfigure}%
\begin{subfigure}{.3333\textwidth}
  \centering
  \includegraphics[width=\linewidth]{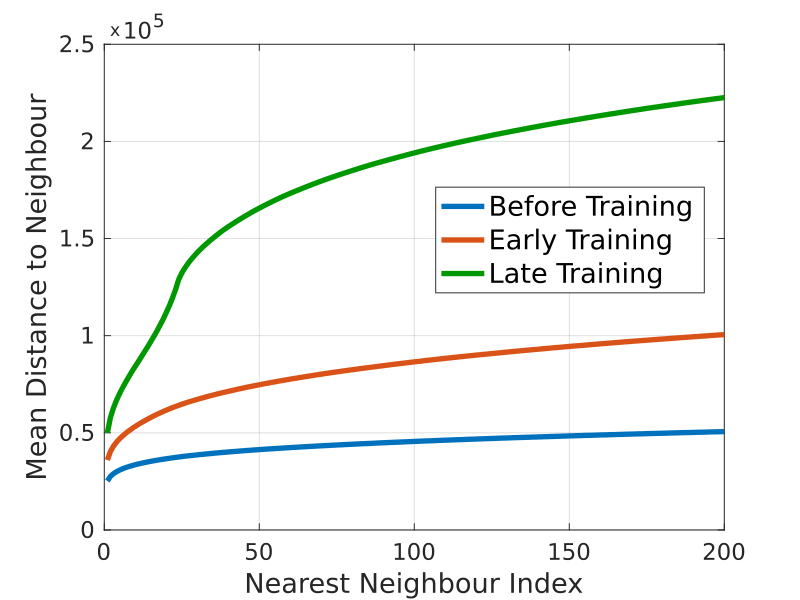}
  \caption{}
  \label{ave_distance}
\end{subfigure}%
\begin{subfigure}{.3333\textwidth}
  \centering
  \includegraphics[width=\linewidth]{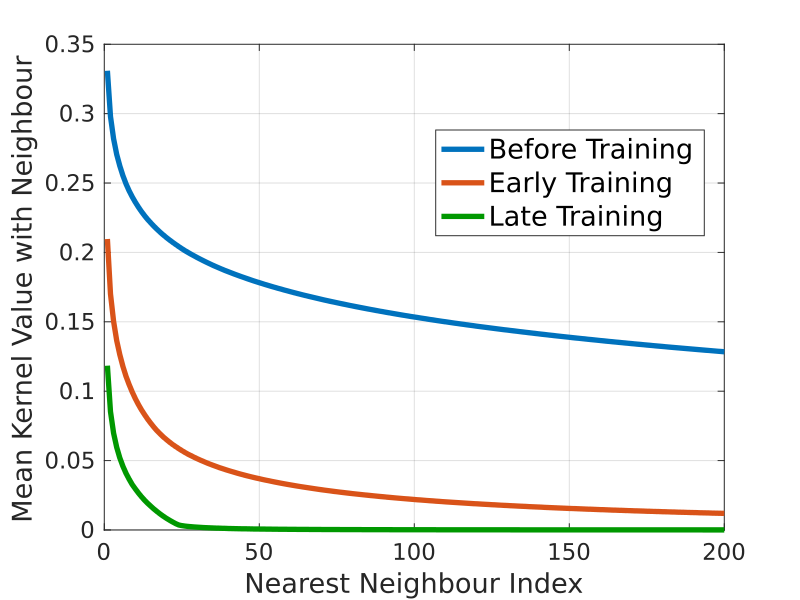}
  \caption{}
  \label{ave_rbf}
\end{subfigure}
\caption{(a) The effect of the number of nearest neighbours considered while training the network. (b) The average distance from training examples to their nearest Gaussian kernel centres, at different points during training. (c) The Gaussian kernel value (Equation \ref{kernel_eq}) between training examples and their nearest kernel centres, at different points during training.}
\label{nn_plots}
\end{figure*}

\section{Supplementary Material}
In this section we provide supplementary qualitative and quantitative results to further evaluate our proposed approach.

\subsection{Ablation Study}
Table \ref{ablation} shows an ablation study for our proposed approach. For each of the five arrangements of settings, the value of the Gaussian kernel $\sigma$ is first tuned to minimise validation loss. The impact of learning the Gaussian kernel weights and fine-tuning network weights is shown.

\subsection{Neighbourhood Size}
Figure \ref{num_nn} shows the impact of the number of nearest neighbours used for each example during training. There is a clear lower bound required for good performance. This is because, as discussed in Section \ref{end_to_end}, the network weights are constantly being updated, but the stored kernel centres are not. As such, we need to consider a larger number of neighbours than if the centres were always up-to-date. Figure \ref{ave_distance} shows the average distance from each training example to its 200 nearest Gaussian kernel centres, at different points during training. Similarly, Figure \ref{ave_rbf} shows the average Gaussian kernel value (from Equation \ref{kernel_eq}) between training examples and their 200 nearest centres. These experiments use a VGG16 architecture.

\subsection{Embedding Space Visualisation}
\begin{figure*}[!t]
	\centering
	\includegraphics[width=0.935\linewidth]{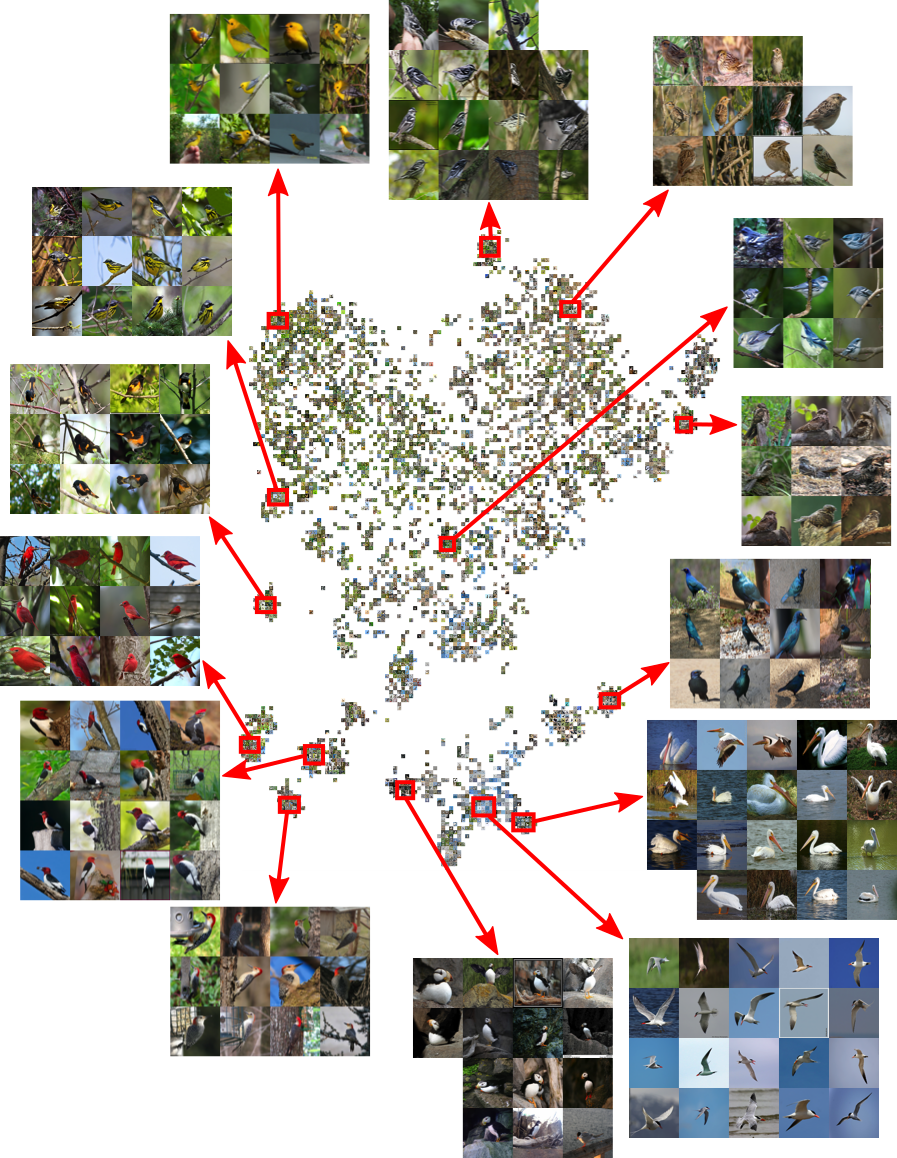}
	\capspace
	\caption{Visualisation of the Birds200 test set embedding space from Section \ref{metric_results}. All species of bird visualised are from withheld classes that were not present during training. Despite this, examples are still well clustered based on species and attributes. The visualisation was obtained using the t-SNE algorithm \cite{maaten2008visualizing}.
	}
	\label{birds_vis}
\end{figure*}

\begin{figure*}[!t]
	\centering
	\includegraphics[width=0.97\linewidth]{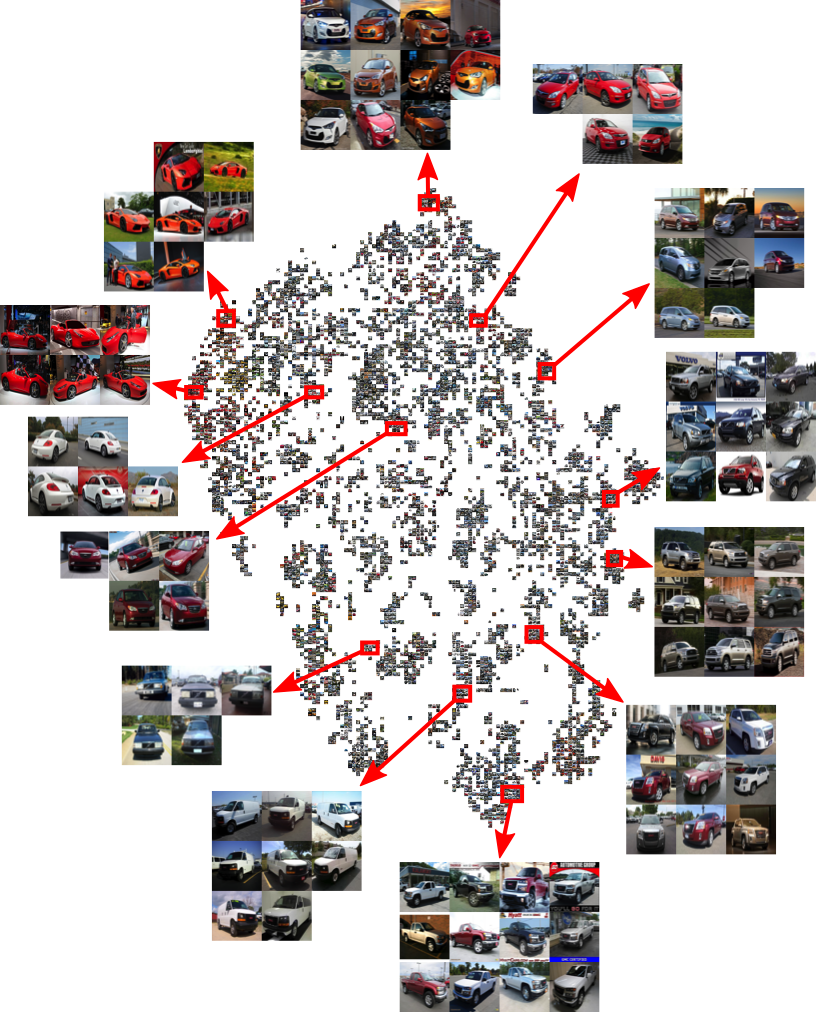}
	\capspace
	\caption{Visualisation of the Cars196 test set embedding space from Section \ref{metric_results}. All models of car visualised are from withheld classes that were not present during training. Despite this, examples are still well clustered based on car model and attributes. The visualisation was obtained using the t-SNE algorithm \cite{maaten2008visualizing}.
	}
	\label{cars_vis}
\end{figure*}

A t-SNE \cite{maaten2008visualizing} visualisation of the learned embedding space for the Birds200 dataset is shown in Figure \ref{birds_vis}. Similarly, Figure \ref{cars_vis} shows a visualisation for the Cars196 dataset. The visualised embeddings are the test set examples from the transfer learning task in Section \ref{metric_results}. The classes shown in the visualisations are withheld and unseen by the network during training. Despite belonging to novel classes, examples are still well clustered based on class and attributes.

\end{document}